\documentclass{article}

\usepackage{arxiv}

\usepackage[utf8]{inputenc} % allow utf-8 input
\usepackage[T1]{fontenc}    % use 8-bit T1 fonts
\usepackage{hyperref}       % hyperlinks
\usepackage{url}            % simple URL typesetting
\usepackage{booktabs}       % professional-quality tables
\usepackage{amsfonts}       % blackboard math symbols
\usepackage{nicefrac}       % compact symbols for 1/2, etc.
\usepackage{microtype}      % microtypography
\usepackage{lipsum}		% Can be removed after putting your text content
\usepackage{graphicx}
\usepackage{natbib}
\usepackage{doi}
\usepackage{verbatim}
\usepackage{subcaption}
\usepackage[misc]{ifsym}

\usepackage{amsmath}
\usepackage{array}
\usepackage{multirow}
\usepackage{subcaption}
\usepackage{listings}

\usepackage{pgf}
\usepackage{tikz}
\usetikzlibrary{arrows,automata}
\usetikzlibrary{positioning}

\tikzset{
    state/.style={
           rectangle,
           rounded corners,
           draw=black, very thick,
           minimum height=2em,
           minimum width=2em,
           inner sep=2pt,
           text centered,
           },
}
\usepackage{placeins}
\usepackage{amssymb}
\usepackage{enumitem}
\usepackage{pifont}

\title{Multilingual Transformers for Product Matching -- Experiments and a New Benchmark in Polish}

\author{ 
    Michał Mo{\.z}d{\.z}onek$^1$,
\href{https://orcid.org/0000-0002-3407-7570}{\includegraphics[scale=0.06]{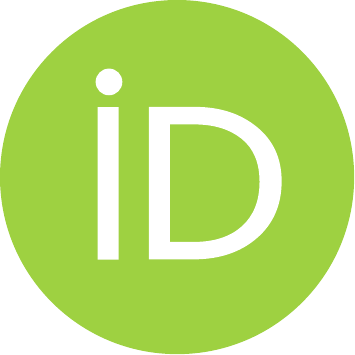}\hspace{1mm}Anna Wr{\'o}blewska$^1$}, \href{https://orcid.org/
0000-0002-3434-6320}{\includegraphics[scale=0.06]{orcid.pdf}\hspace{1mm}Sergiy Tkachuk$^2$}, 
\href{https://orcid.org/0000-0001-6716-610X}{\includegraphics[scale=0.06]{orcid.pdf}\hspace{1mm}Szymon {\L}ukasik$^{2,3}$}\\
 	$^1$Faculty of Mathematics and Information Science, Warsaw University of Technology, Warsaw, Poland \\ Email: \texttt{\{michal.mozdzonek,anna.wroblewska1\}@pw.edu.pl} \\
 	$^2$Systems Research Institute,
		Polish Academy of Sciences\\
		ul.\ Newelska 6, 01-447 Warsaw, Poland\\
		Email: \texttt{\{stkachuk,slukasik\}@ibspan.waw.pl} \\
	$^3$Faculty of Physics and Applied Computer Science,
		AGH University of Science and Technology\\
		al.\ Mickiewicza 30, 30-059 Krak\'{o}w, Poland\\
		Email: \texttt{slukasik@agh.edu.pl} 
 }

\date{}

\renewcommand{\shorttitle}{Multilingual Transformers and a New Benchmark in Polish for Product Matching}

\hypersetup{
pdftitle={Multilingual Transformers for Product Matching -- Experiments and a New Benchmark in Polish},
pdfsubject={cs.CL, cs.LG},
pdfauthor={Michał Mo{\.z}d{\.z}onek, Anna Wr{\'o}blewska, Sergiy Tkachuk, Szymon {\L}ukasik},
pdfkeywords={Product matching, Deep learning, E-commerce, Entity resolution, Multilingual transformers},
}

\begin{document}
\maketitle

\begin{abstract}
Product matching corresponds to the task of matching identical products across different data sources. It typically employs available product features which, apart from being multimodal, i.e., comprised of various data types, might be non-homogeneous and incomplete. The paper shows that pre-trained, multilingual Transformer models, after fine-tuning, are suitable for solving the product matching problem using textual features both in English and Polish languages. We tested multilingual mBERT and XLM-RoBERTa models in English on Web Data Commons - training dataset and gold standard for large-scale product matching. The obtained results show that these models perform similarly to the latest solutions tested on this set, and in some cases, the results were even better. 

Additionally, we prepared a new dataset -- ProductMatch.pl -- that is entirely in Polish and based on offers in selected categories obtained from several online stores for the research purpose. It is the first open dataset for product matching tasks in Polish, which allows comparing the effectiveness of the pre-trained models. Thus, we also showed the baseline results obtained by the fine-tuned mBERT and XLM-RoBERTa models on the Polish datasets.
\end{abstract}


\section{Introduction}

\begin{figure}[!htb]
    \centering
    \includegraphics[width=0.5\textwidth]{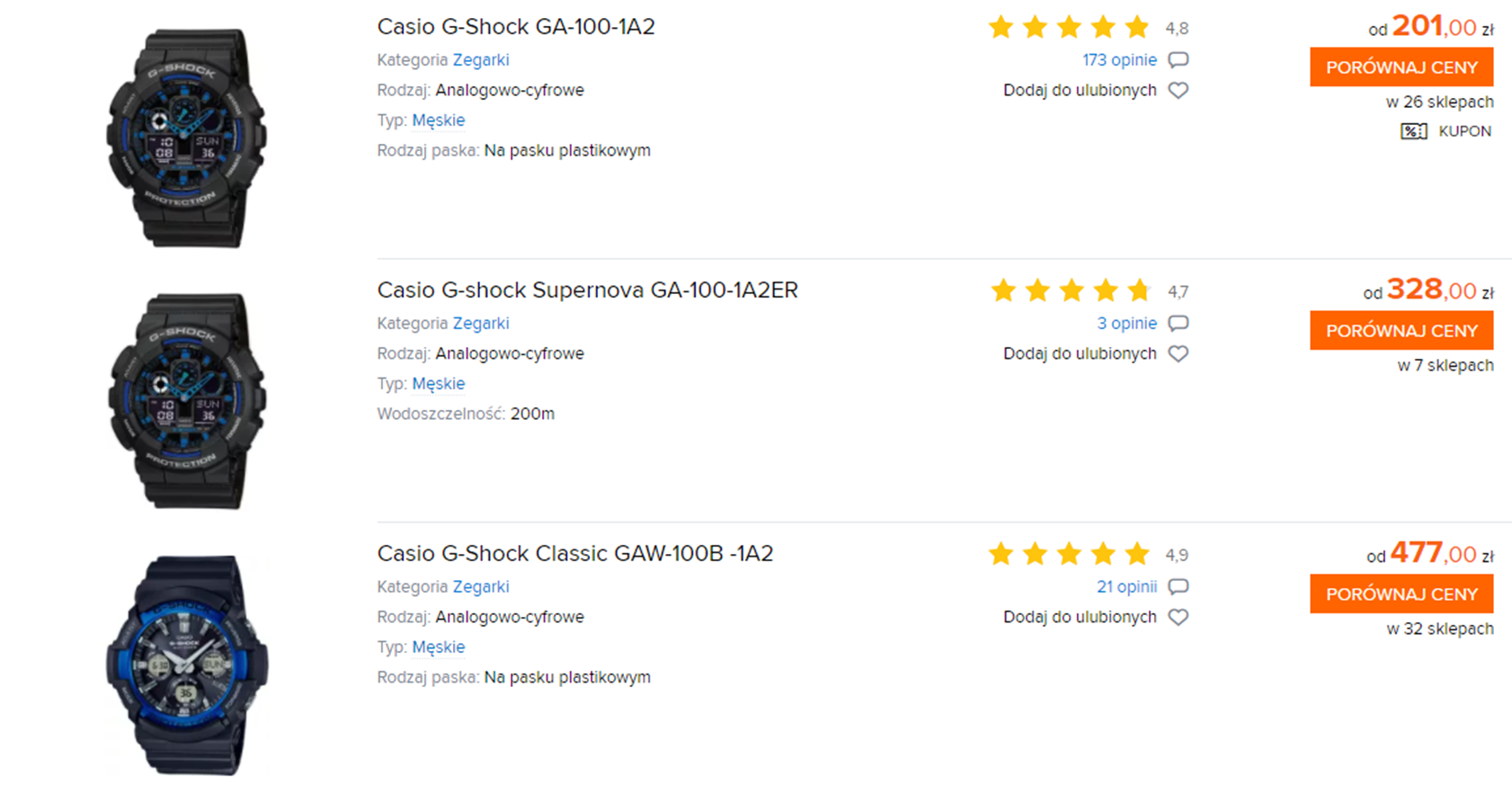}
    \caption[Offers grouping error example]
    {Offers grouping error example. Offers of the same watch are split into two groups. This issue is likely due to some sellers using a abbreviated and some using a full product title. One group has a significantly lower price and for that reason seems more attractive. Source: www.ceneo.pl\label{fig:ceneo_error}}
\end{figure}

Moving from traditional to Internet stores led to the global availability of offers. However, the plethora of offers means that buyers are often overwhelmed by their multitude and cannot choose the offer that suits their preferences best. Thus, many services are dedicated to aggregate offers to overcome this challenge of various descriptions being provided for offers of the same product. For instance, in Poland, one of such services is \textit{ceneo.pl}, whereas \textit{shopping.com} constitutes an example from the US market. Aggregation sites collect listings for the same product and represent it by a common title, image, and optional attributes within such a group. Figure~\ref{fig:ceneo_error} presents an example from \textit{ceneo.pl}.

It turns out that the process of offers grouping is complex, and~potential product misplacement causes considerable repercussions. Figure~\ref{fig:ceneo_error} provides an example of this problem and its consequence. In literature, this issue is related to the Product Matching Task (PM). Product Matching is a problem of identifying and linking/grouping different manifestations (shop offers) of the same real-world product.

The latest research \cite{Mudgal2018,ditto} on the Product Matching problem (and more general Entity Resolution problem, where we are not restricted only to products) shows that deep neural networks accomplish the best results. In work~\cite{ditto}, the authors showed that transfer learning could be applied to fine-tune Transformer models to solve this problem. However, this research was conducted only for datasets in English.

English is the most widely used language globally, but many nations have other mother tongues. Each of these countries has its local e-commerce market operating in its specific language, so it is worth considering solving the PM problem in this context. Multilingual models of the Transformer type give us such a possibility. As the Ditto~\cite{ditto} project showed, models such as the BERT~\cite{devlin2019bert} perform well for the PM task. However, how Transformers will perform on data in other languages, particularly in Polish, has not been investigated. It may be interesting to see if using the same pre-trained model to solve the PM problem in English and Polish is viable.

Consequently, this paper's contribution is as follows:
\begin{itemize}
    \item Testing and demonstrating that transfer learning can be used for languages other than English (especially Polish) in the frame of the product matching task,
    \item Providing state-of-the-art level results for fine-tuned multilingual BERT and XLM-RoBERTa models trained on English WDC dataset,
    \item Preparing Polish datasets for product matching problem with model baselines.\footnote{Our code for training and testing models, and the Polish dataset is available at our GitHub repository: \url{https://github.com/grant-TraDA/mlt4pm}.}
\end{itemize}

The following sections describe a solution to the product matching problem in Polish and English languages. In Section~\ref{s:releatedWork}, we discuss well-known solutions to both Entity Resolution and Product Matching tasks. Section~\ref{s:wdc} presents the most important dataset for model evaluation on the Product Matching problem (Web Data Commons). Section~\ref{s:polishDataset}  describes the creation of the Polish product matching dataset. Section~\ref{s:methodology} contains a description of the testing procedure. Section~\ref{s:results} gathers our results obtained by models trained on the Polish and English datasets. In terms of the English datasets, we compare results to other reported solutions. Conclusion, given in Section~\ref{s:conclusions}, summarizes our work and highlight opportunities for further development of research in this field.

\section{Related Work} \label{s:releatedWork}

Solutions developed so far within the scope of this research could be divided into more general -- related to Entity Resolution (ER) problems -- and specific to product matching tasks. Entity resolution is well-known in the literature and described in the book~\cite{christen2012data}. Many approaches to this problem have been proposed so far, ranging from probabilistic models~\cite{FellegiIvanP1969ATfR,ERMarkovLogic}, algorithms based on rigid rules~\cite{LingliLi2015RMfE}, and ending with machine learning models~\cite{peterKNN2008}. In 2018, Mudgal~\cite{Mudgal2018} compared the effectiveness of deep learning methods and the Magellan system.\footnote{Magellan system description available at \cite{Magellan}, project website -- \href{https://sites.google.com/site/anhaidgroup/current-projects/magellan}{sites.google.com/site/anhaidgroup/current-projects/magellan}} The Magellan system at that time was considered the state-of-the-art solution. The authors showed that deep learning models perform better in this task. A considerable difference was visible when offers were presented in the form of text or noisy data.\footnote{The authors referred to the term noisy data when some attributes are contained within the other attributes. For example, the producer's name is part of the title, and the appropriate corresponding field is left blank.} As the result of this research, the Deepmatcher system was created, which is available as a Python package.\footnote{\href{https://github.com/anhaidgroup/deepmatcher}{github.com/anhaidgroup/deepmatcher}}

The team preparing the WDC set (see Section~\ref{s:wdc}) tested several known algorithms such as TF-IDF with cosine distance, Magellan system, Random Forest, XGBoost, co-occurrence matrix (see~\cite{WDCwwwv2} for detail) with SVM and DeepMatcher system. 
Table~\ref{tab:wdc_old} presents F1 scores for each method. 
The DeepMatcher system using RNN networks did the best among all tested algorithms. The obtained values confirm the results of Mudgal's~\cite{Mudgal2018} work, which concerned the more general ER problem. The studies demonstrate the superiority of deep learning methods for the PM task.

Another solution to the ER problem presented in September 2020 is the Ditto system~\cite{ditto}. The key element of the whole system is the matching module. The module takes two entities as an input and returns information if these entities refer to the same real-world product (output $=1$) or not (output $=0$). Different Transformer models were tested as a classifying model, including: DistilBERT~\cite{sanh2020distilbert}, BERT~\cite{devlin2019bert}, RoBERTa~\cite{liu2019roberta} and XLNet~\cite{yang2020xlnet}. All models were pre-trained on larger datasets and made available under the Transformers library~\cite{wolf2020huggingfaces}. 
%Figure~\ref{fig:dittoModel} presents the schematic design of the matching model.
	
The authors performed tests on WDC sets (see Section~\ref{s:wdc}) and on datasets mentioned in works on DeepMatcher~\cite{Mudgal2018}. The comparison models were, among others, the previously mentioned Magellan and DeepMatcher systems. The Ditto system achieved better results than the models discussed above. 
All the methods are compared in Table~\ref{tab:wdc_old}. 
\footnote{The project is publicly available as a repository: \url{https://github.com/megagonlabs/ditto}}

\begin{table*}[!htb]
    \caption{F1 scores for models trained on English WDC datasets. Except for Ditto, all results are reported in~\cite{WDCwwwv2}, and Ditto is reported in~\cite{ditto}}.
   \label{tab:wdc_old}
	\centering
	\begin{tabular}{ccccccc}
	\toprule
		dataset & dataset & TF-IDF & \multirow{2}{*}{Magellan} & \multirow{2}{*}{Co-Occ} & \multirow{2}{*}{Deepmatcher} & \multirow{2}{*}{Ditto} \\
		type & size & cosine & & & & \\
	\midrule
	\multirow{4}{*}{Cameras} 	& small  & 64 & 55.63 & 61.75 & 68.59 & \textbf{80.89} \\
							 	& medium & 64 & 59.44 & 70.26 & 76.53 & \textbf{88.09} \\
							 	& large  & 64 & 60.76 & 82.08 & 87.19 & \textbf{91.23} \\
							 	& xlarge & 64 & 62.84 & 72.12 & 89.21 & \textbf{93.78} \\
	\midrule
	\multirow{4}{*}{Computers} 	& small  & 57 & 59.23 & 60.99 & 70.55 & \textbf{80.76} \\
							   	& medium & 57 & 64.13 & 74.55 & 77.82 & \textbf{88.62} \\
							 	& large  & 57 & 69.22 & 81.51 & 89.55 & \textbf{91.70} \\
							 	& xlarge & 57 & 68.65 & 83.59 & 90.80 & \textbf{95.45} \\
	\midrule
	\multirow{4}{*}{Shoes} 		& small  & 57 & 64.07 & 70.69 & 73.86 & \textbf{75.89} \\
							 	& medium & 57 & 62.96 & 79.19 & 79.48 & \textbf{82.66} \\
							 	& large  & 57 & 63.45 & 71.70 & \textbf{90.39} & 88.07 \\
							 	& xlarge & 57 & 63.62 & 69.32 & \textbf{92.61} & 90.10 \\
	\midrule
	\multirow{4}{*}{Watches} 	& small  & 63 & 67.16 & 62.88 & 66.32 & \textbf{85.12} \\
							 	& medium & 63 & 65.85 & 69.54 & 79.31 & \textbf{91.12} \\
							 	& large  & 63 & 68.27 & 78.14 & 91.28 & \textbf{95.69} \\
							 	& xlarge & 63 & 60.20 & 77.61 & 93.45 & \textbf{96.53} \\
	\bottomrule
	\end{tabular}
\end{table*}

Similar challenges were also recently addressed by~\cite{peeters2021cross, peeters2022cross,peeters2022contr} in the frame of the cross-language models' studies. The authors emphasize enrichment of the non-English data and benefits of using Transformers architecture in the PM tasks for low-resource languages using German as an example.

\section{English Dataset -- Web Data Commons} \label{s:wdc}

Due to the high level of specialization in combining offers, a small number of datasets in the public domain allow for training and evaluation of created solutions. The most extensive, widely shared collection is ''Web Data Commons -- Training Dataset and Gold Standard for Large-Scale Product Matching''\footnote{\href{http://webdatacommons.org/largescaleproductcorpus/v2/index.html}{webdatacommons.org}} (WDC for short) prepared by the staff of the University of Mannheim. The first version of the set is described in~\cite{WDCarticle}, the second -- in~\cite{WDCschemaorg}. The authors of these studies extracted offers from websites collected under the CommonCrawl project.\footnote{\href{http://commoncrawl.org/}{commoncrawl.org}} The selection of offers was based on the schema.org\footnote{\href{https://schema.org/}{schema.org}} tags, which Internet sellers had started using in the last five years~\cite{WDCarticle}. The authors cleaned and simplified the record structure in the second version of the collection. The collection contains 26 million offers allocated to 16 million groups, of which 1.1~million are groups containing three or more offers. A subset of offers in English only, with 16 million entries, was separated from the dataset. The offers in this collection related to different categories -- computers, cameras, watches, and shoes -- were separated. These are all categories available as open preprocessed datasets. All offers in these collections have a title. Smaller datasets were also prepared, known as the "Gold Standard," containing 300 pairs of corresponding offers and 800 pairs of non-corresponding offers. The pairs included in the "Gold Standard" collection were verified manually. According to~\cite{WDCschemaorg}, ''Gold Standard'' should be used as a test dataset for model evaluation.

The training datasets are available in different sizes, varying from small to extra large. In every dataset, the ratio between positive and negative pairs is 1:3. Table~\ref{tab:wdc_sizes} presents the exact sizes for each dataset. The proportions between the different collections within one category are as follows: 1 -- small, 3 -- medium, 15 -- large, 50 -- extra-large (xlarge).

\begin{table}[!hpt]
	\caption{WDC datasets sizes. Source:~\cite{WDCwwwv2}}
	\label{tab:wdc_sizes}
	\centering
	\begin{tabular}{ccccc}
	\toprule
		Category & Size & Positive & Negative & Total \\
	\midrule
		\multirow{4}{*}{Cameras}   & Small  &   486 &  1,400 &  1,886 \\
								   & Medium & 1,108 &  4,147 &  5,255 \\
								   & Large  & 3,843 & 16,193 & 20,036 \\
								   & xLarge & 7,178 & 35,099 & 42,277 \\
	\midrule
		\multirow{4}{*}{Computers} & Small  &   722 &  2,112 &  2,834 \\
								   & Medium & 1,762 &  6,332 &  8,094 \\
								   & Large  & 6,146 & 27,213 & 33,359 \\
								   & xLarge & 9,690 & 58,771 & 68,461 \\
	\midrule
		\multirow{4}{*}{Watches}   & Small  &   580 &  1,675 &  2,255 \\
								   & Medium & 1,418 &  4,995 &  6,413 \\
								   & Large  & 5,163 & 21,864 & 27,027 \\
								   & xLarge & 9,264 & 52,305 & 61,569 \\
	\midrule
		\multirow{4}{*}{Shoes}     & Small  &   530 &  1,533 &  2,063 \\
								   & Medium & 1,214 &  4,591 &  5,805 \\
								   & Large  & 3,482 & 19,507 & 22,989 \\
								   & xLarge & 4,141 & 38,288 & 42,429 \\
		
	\bottomrule		
	\end{tabular}
\end{table}

\section{Our Open Polish Dataset} \label{s:polishDataset}

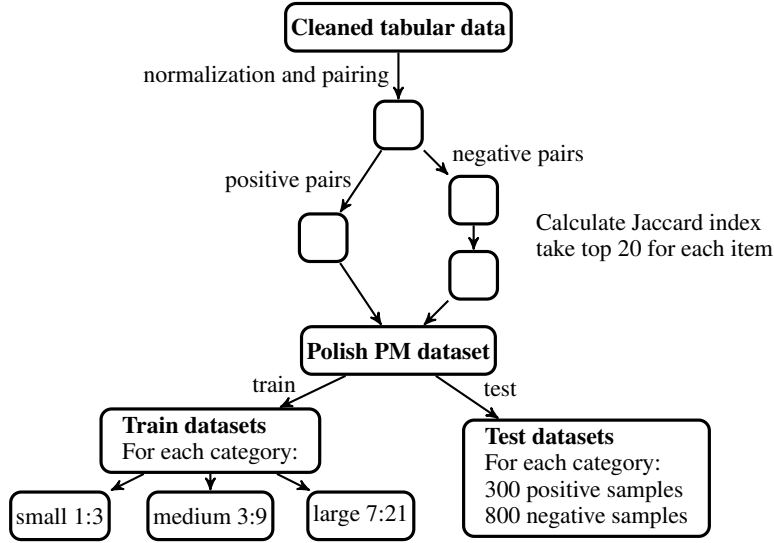
\begin{figure*}[hpt]
	\centering
	\begin{tikzpicture}[
	->,>=stealth',
	thick,
	scale=0.5
	]
    \small
 \node[state,
 		anchor=center
 ] (Cleaned) {
 	\textbf{Cleaned tabular data} 
 	};
 \node[state, 
 		below of=Cleaned,
 		node distance=1.3cm,
 		anchor=center
 ] (Pairs) {};
 
 \node[state, 
 		below of=Pairs,
 		xshift=1cm,
 		node distance=1cm
 ] (Neg) {};
 \node[state, 
 		below of=Pairs,
 		xshift=-1cm,
 		node distance=1.5cm
 ] (Pos) {};
 
 \node[state, 
 		below of=Neg,
 		node distance=1cm
 ] (Neg20) {};
 
 \node[state, 
 		below of=Pairs,
 		node distance=3cm
 ] (PairedDataset) {\textbf{Polish PM dataset}};
 
  \node[state, 
 		below of=PairedDataset,
 		node distance=1.7cm,
 		xshift=2.5cm
 ] (Test) {
	 \begin{tabular}{l}
	\textbf{Test datasets}\\
	For each category:\\ 
	300 positive samples \\ 
	800 negative samples
	\end{tabular}
	};
	
	 \node[state, 
 		below of=PairedDataset,
 		node distance=1.2cm,
 		xshift=-2.5cm
 ] (Train) {
 	\begin{tabular}{l}
	\textbf{Train datasets}\\
	For each category:\\
	\end{tabular}
 };
 
 	\node[state, 
 		below of=Train,
 		node distance=1cm,
 		xshift=-2cm
 ] (Small) {small 1:3};

 \node[state, 
 		below of=Train,
 		node distance=1cm
 ] (Medium) {medium 3:9};
 
  	\node[state, 
 		below of=Train,
 		node distance=1cm,
 		xshift=2cm
 ] (Large) {large 7:21};
 
 \path (Cleaned) 	edge  node[left]{normalization and pairing} (Pairs)
	(Pairs) 	edge  node[left]{positive pairs} (Pos)
	(Pairs) 	edge  node[right, xshift=0.1cm, yshift=0.1cm]{negative pairs} (Neg)
	(Neg) 	edge  node[right, xshift=0.5cm]{
		\begin{tabular}{l}
		Calculate Jaccard index \\ 
		take top 20 for each item
		\end{tabular}
	} (Neg20)
	(Pos) 	edge  node[right]{} (PairedDataset)
	(Neg20) 	edge  node[right]{} (PairedDataset)
	(PairedDataset) 	edge  node[right, xshift=0.1cm, yshift=0.1cm]{test} (Test)
	(PairedDataset) 	edge  node[left, xshift=-0.1cm, yshift=0.1cm]{train} (Train)
	(Train) 	edge  node[right]{} (Small)
	(Train) 	edge  node[right]{} (Medium)
	(Train) 	edge  node[right]{} (Large)
 ;

	\end{tikzpicture}
	\caption[Creating the Polish PM datasets]
	{The process of creating the Polish PM datasets. In each training set, the ratio of positive to negative samples is 1:3.}
	\label{fig:creatingPolishPM}
\end{figure*}

For the Polish product matching dataset, the data were collected from popular Polish stores. Then the data were cleaned, reduced to tabular form, and then converted to a format compatible with the WDC dataset (see Section~\ref{s:wdc}). Figure~\ref{fig:creatingPolishPM} presents the general schema of the process of creating our Polish dataset.

The procedure of cleaning was as follows. Firstly, the records with missing seller details were deleted, and only the records containing both EAN and title columns remained. The EAN column uniquely identifies the product and is required in the offers pairing phase. While most available solutions use either the title or combined with other attributes, all records in the WDC dataset must contain a title. 

In the next step, we extracted the main category from the category tree and distinguished the two largest categories from the data: drinks (pl. "napoje") and household chemistry (pl. "chemia"). The name unification also played a significant role as stores used different names for the same category. Then we combined the EAN and seller columns and removed duplicates due to this new concatenated column. This step was crucial because the sellers often displayed the same product several times on their website. The next step was to select only those offers of products available in at least two stores. This way, offers were obtained to build a dataset compliant with the WDC standards. Finally, the column names were converted to the names appearing in the WDC standard -- the resulting dataset achieved still a non-standardized naming convention.\footnote{The dataset in this form is at our GitHub: \textit{pl\_wdc\_non\_normalized.json.gz}.} 
A second normalized version of the set was also prepared: upper-case letters were lower-cased, non-alphanumeric characters, and white characters appearing many times in sequences were removed.\footnote{The dataset in this form is at our GitHub:  \textit{pl\_wdc\_normalized.json.gz}.} 

Next, offers were combined into pairs to build training and testing sets for the PM problem. Within each category, the Cartesian product of offers was created (every offer was paired with all offers in the category). \textit{\_left}, and \textit{\_right} suffixes were added to column names to distinguish offers. The pairs where identical EAN (European Article Number) of the first and second offer are called positive pairs. The final dataset added that information as a target column (label). The other offer pairs were negative. Due to the small number of positive pairs (around 3,600 for household chemistry and 2,700 for drinks), all such pairs were kept for further processing. It is the first difference with the WDC format; in the WDC dataset, only a subset of positive pairs was taken into the final datasets. For negative pairs, the Jaccard index was computed for the title columns. The most similar -- 20 highest-scoring offers were then selected for each product. In this way, negative pairs were obtained, i.e., those in which the offers refer to different products but at the same time have similar titles. Choosing negative pairs at random might produce elementary samples, which would not be useful in training.
	
Three hundred positive and 800 negative pairs were selected from the training dataset and created the testing dataset. Three sizes of training sets were prepared with the following proportions: small (1), medium (3), and large (7). In each set, the ratio of positive to negative samples is 1:3. The dataset size and distribution are mostly consistent with the WDC datasets. However, contrary to the WDC project, large datasets have a size ratio to the small datasets of 7:1, wherein WDC a size ratio is 15:1. This change was caused by a low number of positive samples in the Polish datasets. Moreover, the pairs in test datasets were not validated manually but prepared according to the above process. Table~\ref{tab:pl_pm_sizes} presents the exact sizes for each Polish dataset.
	
	\begin{table}[hpt]
    	\caption{Polish PM datasets sizes.}
		\label{tab:pl_pm_sizes}
		\centering
		\begin{tabular}{ccrrr}
		\toprule
			Category & Size & Positive & Negative & Total \\
		\midrule
			\multirow{3}{*}{Household chemistry}    & Small  &   503 &  1,509 &  2,012 \\
									   & Medium & 1,509 &  4,527 &  6,036 \\
			(pl. chemia)		   & Large  & 3,521 & 10,563 & 14,084 \\
		\midrule
			\multirow{3}{*}{Drinks}    & Small  &   381 &  1,143 &  1,524 \\
									   & Medium & 1,143 &  3,429 &  4,572 \\
			(pl. napoje)			   & Large  & 2,667 &  8,001 & 10,668 \\
		\midrule
			\multirow{3}{*}{All}       & Small  &   884 &  2,652 &  3,536 \\
									   & Medium & 2,652 &  7,956 & 10,608 \\
									   & Large  & 6,188 & 18,564 & 24,752 \\
		\bottomrule		
		\end{tabular}
	\end{table}

\section{Experimental Procedure} \label{s:methodology}
Our goal is to test the effectiveness of multilingual Transformers models in the product matching task on WDC and our Polish dataset.

Models were implemented with the HuggingFace Transformers library.\footnote{\url{https://huggingface.co/transformers/}}~\cite{wolf2020huggingfaces} We used sequence classification layers with two output classes as our goal is to distinguish corresponding and non-corresponding pairs of offers. Two types of pre-trained models were used: mBERT~\cite{devlin2019bert} and XLM-RoBERTa~\cite{xlm_roberta}. Table~\ref{tab:model_spec} presents the specification for each model.

    \begin{table}[hpt]
        \caption{Models' specification. Note: Num languages means the number of languages, Num parameters -- the number of parameters.}
        \label{tab:model_spec}
        \centering
        \begin{tabular}{ccc}
        \toprule
            Model & Num languages & Num parameters  \\
        \midrule
             bert-base-multilingual-uncased\footnote{\url{https://github.com/google-research/bert/blob/master/multilingual.md}} & 102 & 110M  \\
             xlm-roberta-base\footnote{\url{https://github.com/pytorch/fairseq/tree/master/examples/xlmr}} & 100 & 250M  \\
        \bottomrule
        \end{tabular}
    \end{table}
	
We carried out: (1) tests with fixed random seed (42) for each dataset to make the whole experiment reproducible and (2) experiments with random seed to calculate scores standard error. For the experiments, we used datasets in English, provided under the WDC project (see Section~\ref{s:wdc}) and our dataset in Polish (see Section~\ref{s:polishDataset}). Each dataset was divided into a training set and a validation set using the test\_train\_split function\footnote{scikit-learn test\_train\_split documentation  -- \href{https://scikit-learn.org/stable/modules/generated/sklearn.model_selection.train_test_split.html}{link}} from the scikit-learn package. The validation set size was set to 20\% and the random\_state parameter to 42 (random in the second group of experiments). The validation set is used in the training process to select the best model instance. The gold standard sets (see Section~\ref{s:wdc}) were used as a test set for the WDC set. For the datasets in Polish, the previously prepared test datasets were utilized in a test phase.
	
Preparing the data for the model, first of all, consisted of selecting the columns to be used. In the article related to the Ditto~\cite{ditto} system, the best results were obtained where only the title column was used. Our solution, the same as Ditto~\cite{ditto}, utilizes the Transformers model. For this reason, we use only the title column in our experiments. The data were then concatenated in the following order:
    	\begin{enumerate}
    		\item token [CLS]  -- needed for the classification task,
    		\item data of the left offer,
    		\item token [SEP] -- separator,
    		\item data of the right offer,
    		\item token [SEP] -- marking the end.
    	\end{enumerate}
Thus, we get a single input string for a model. In the next step, the tokenizer (provided with the pre-trained model) transforms the input string into a list of tokens. Data in that preprocessed form was ready for training. The Trainer class provided as part of the Transformers library in the training phase was employed. Table~\ref{tab:trainParams} presents the training parameters that were set for all models. 
	
	\begin{table}[!hpt]
	    \caption{Training parameters.}
		\label{tab:trainParams}
		\centering
		\addtolength{\tabcolsep}{3pt}  
		\begin{tabular}{rl}
		\toprule
			Parameter & Value \\
		\midrule
			num\_train\_epochs & 10\\
			train\_batch\_size & 16\\
			eval\_batch\_size & 64\\
		\midrule	
			optimizer & AdamW (default) \\
			learning\_rate (initial) & 5e-5 (default)\\
			adam\_beta1 & 0.9 (default)\\
			adam\_beta2 & 0.999 (default)\\
			adam\_epsilon & 1e-8 (default)\\
			weight\_decay & 0.01\\
			warmup\_steps & $\frac{\text{len(train\_dataset)} \times \text{num\_train\_epochs}}{2 \times \text{train\_batch\_size}}$\\
			fp16 & True \\
			seed & 42 (default) / random\\
		\midrule
			logging\_steps & 10	\\
			evaluation\_strategy & epoch \\
			load\_best\_model\_at\_end & True \\
			metric\_for\_best\_model & eval\_f1 \\
			save\_total\_limit & 5 \\
		\bottomrule
		\end{tabular}
		\addtolength{\tabcolsep}{-3pt}
	\end{table}
	
%   \begin{figure}[hpt]
% 		\centering
% 		\includegraphics[width=0.3\textwidth]{images/train_learning_rate.png}
%         \caption{Learning rate scheduling used in training. In the first half, the learning rate is increased linearly from $\approx 1e-7$ to $\approx 5e-5$. In the second - decreased to its initial value.}
% 		\label{fig:learning_rate}
% 	\end{figure}

AdamW~\cite{AdamW} was used as an optimizer, as it is the default optimizer in the Transformer library. The learning rate for the first half of the training was increased linearly from $\approx 1e-7$ to $\approx 5e-5$ and then decreased linearly back to the initial value. This approach was inspired by the 1cycle~\cite{1cycle} method proposed by Leslie Smith. To speed up the training process, the mixed half-precision floating-point acceleration (fp16) was used~\cite{fp16}. The number of epochs was pre-set to 10. It is a compromise between the quality of the solution and the time needed to perform the test. After each epoch, the model is tested on a validation set, and the accuracy, precision, recall, and F1 metrics are calculated. Then the model checkpoint is saved. Due to space availability, the number of saved checkpoints was limited to 5. After training, the checkpoint with the best result in the F1 metric is selected, and the saved parameter weights are loaded to the model. The test set is used to calculate the final result given metrics: accuracy, precision, recall, and F1 score.
    
After all experiments, the mean values for F1 score, precision, and recall were calculated. In addition, the standard error was added to the final results. Formula~\ref{eq:std_err} presents how it was calculated. The $\hat{\sigma}$ is the standard error, and the $\sigma$ is the standard deviation. The $t_{ppf}$ is the percent point function of t-Student distribution (inverse of cumulative distribution function -- percentiles). Confidence level was set to $95\%$ (term \textit{conf} in Formula~\ref{eq:std_err}). The used confidence interval was two-sided.
    
    \begin{equation}
		\hat{\sigma} = t_{ppf}\left( \frac{1+\mbox{conf}}{2}, n-1 \right) \times \sigma \label{eq:std_err}
	\end{equation}

\section{Results}  \label{s:results}
    \newcommand{\resinfo}{Mean value and standardized error (confidence level 95\%) for each dataset were calculated from \textbf{4} samples. For further information on how standardized error was calculated see Section~\ref{s:methodology}.}

\subsection{Polish Dataset}  \label{s:results_pl}

First, we tested the mBERT model on the Polish dataset. The model was loaded from the Transformers library model hub.\footnote{Transformers model hub -- \href{https://huggingface.co/models}{link}} The exact name of the model checkpoint was ''bert-base-multilingual-uncased''. As the name suggests, this is a multilingual model, as it was pre-trained on Wikipedia articles in 102 languages. The suffix "uncased" means the model does not distinguish between uppercase and lowercase letters. We fine-tune the mBERT model on all available Polish datasets in all sizes. Next, we checked how the pre-trained XLM-RoBERT model could handle the same task. We used the ''xlm-roberta-base'' checkpoint for model initialization. The same parameters were used as in the case of training the mBERT model. 

The obtained results were used to create a baseline for the data set in Polish. It is the first attempt to evaluate the performance of models on this dataset. 

We used the F1 score as a metric to compare the models. Table~\ref{tab:baseline} and Figure~\ref{fig:results-polish} present the results. The mBERT model achieved better results on the small and medium datasets beating XLM-RoBERTa in all three categories. On the other hand, on large datasets, XLM-RoBERTa works slightly better. Only on \textit{''Chemia'' (eng. household chemistry)  large} the mBERT achieved better results. The size of the model might explain it. The mBERT has fewer parameters ($\approx 168M$) than the XLM-RoBERTa ($\approx 270M$) and should fit data faster. Nevertheless, differences between mBERT and XLM-RoBERTa are not significant in most cases.

    \begin{table}
        \caption{F1 scores for mBERT and XLM-RoBERTa trained on Polish datasets. \resinfo}
    	\label{tab:baseline}
    	\centering
    	\begin{tabular}{cccc}
    	\toprule
    		dataset type & dataset size & mBERT & XLM-RoBERTa \\
    	\midrule
    	\multirow{2}{*}{Household chemistry} & small  & $\mathbf{85.73(\pm 1.89)}$ & $83.15(\pm 4.15)$ \\
    							& medium & $\mathbf{90.78(\pm 3.03)}$ & $89.03(\pm 5.96)$ \\
    	(pl. chemia)		& large  & $\mathbf{93.25(\pm 1.77)}$ & $92.52(\pm 1.77)$ \\
    	\midrule
    	\multirow{2}{*}{Drinks} & small  & $\mathbf{85.17(\pm 1.61)}$ & $84.43(\pm 7.16)$ \\
    							& medium & $\mathbf{88.98(\pm 2.63)}$ & $88.44(\pm 2.88)$ \\
    	(pl. napoje)		    & large  & $89.39(\pm 2.12)$          & $\mathbf{89.93(\pm 3.99)}$ \\
		\midrule	    							
    	\multirow{3}{*}{All} 	& small  & $\mathbf{85.73(\pm 1.96)}$ & $84.67(\pm 9.03)$ \\
    							& medium & $\mathbf{90.78(\pm 1.13)}$ & $88.63(\pm 2.79)$ \\
    							& large  & $91.41(\pm 3.17)$          & $\mathbf{91.61(\pm 1.39)}$ \\
    	\bottomrule
    	\end{tabular}
    \end{table}
    
\subsection{WDC Dataset} \label{s:results_wdc}

In the next phase, we used the WDC dataset (see Section~\ref{s:wdc}) as the second dataset for testing. We loaded the pre-trained mBERT and XLM-RoBERTa models from the same checkpoints as in the tests on Polish datasets (see Section~\ref{s:results_pl}). We rejected the files of the ''All'' type and files of the ''xlarge'' size from the available datasets. The given datasets are much larger than others, so the learning time of the models was too long for the available environment. 
	    
\begin{figure*}[!htb]
    \centering
    \begin{minipage}[t]{.48\textwidth}
        \centering
        \includegraphics[width=\linewidth]{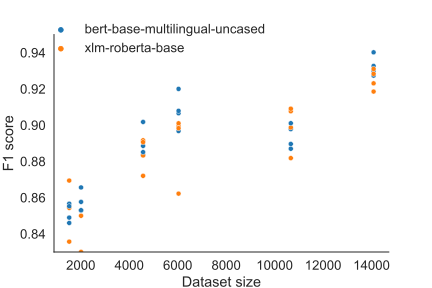}
        \caption[F1 score vs dataset size for models trained on Polish datasets]
   			{F1 score vs dataset size for models trained on Polish datasets.}
        \label{fig:results-polish}
    \end{minipage}%
    \hfill
    \begin{minipage}[t]{.48\textwidth}
        \centering
        \includegraphics[width=\linewidth]{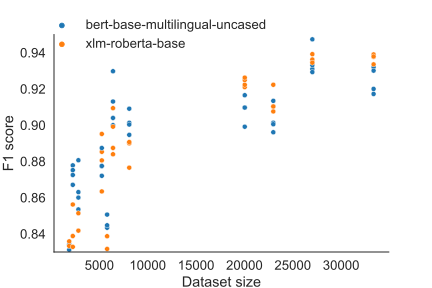}
        \caption[F1 score vs dataset size for models trained on WDC datasets]
   			{F1 score vs dataset size for models trained on WDC datasets.}
        \label{fig:results-wdc}
    \end{minipage}
\end{figure*}

We compared the obtained results of the mBERT and XLM-RoBERT models with the results presented in the paper describing the Ditto~\cite{ditto} system. For the comparison, we also considered the results obtained by the DeepMatcher~\cite{Mudgal2018} model available in the baseline section of the WDC~\cite{WDCwwwv2} project website. The other models listed on the WDC website~\cite{WDCwwwv2} scored much lower F1 scores than our solution, and it makes no sense to compare them with the results obtained in this project. Table~\ref{tab:wdc_comparison} presents a comprehensive list with the best results highlighted for each dataset (see also Figure~\ref{fig:results-wdc}). The results for the mBERT and XLM-RoBERT models on xlarge datasets were not obtained and are marked with ''-''. 
	    
The mBERT and XLM-RoBERT models did very well and, in many cases, obtained better results than those presented in other studies (see Table~\ref{tab:wdc_comparison}). Again, we can observe that the mBERT model receives better results than the other models, especially on sets of smaller sizes (small, medium). We deal with similar results in the set in Polish (see Section~\ref{s:results_pl}), where only the mBERT and XLM-RoBERT models are compared. The Deepmatcher project only wins once in the 'Shoes xlarge' dataset. All collections from the ''Shoes'' group stand out because the results obtained are several percent worse than the other collections (e.g., for the mBERT model, the set small F1 score -- 79.20 vs the others small -- 82.13, 86.43, 87.31). The Ditto project gets the best results in the other charts for xlarge size. The XLM-RoBERT model worked exceptionally well on ''large'' sets -- only the Ditto project outperforms it in the case of ''Watches large'' sets.
	    
The obtained results show that multilingual models can be used to solve the product matching problem. Moreover, their effectiveness does not differ from the results presented in other works, and sometimes it is even better.
	      
    \begin{table*}
        \caption{F1 scores for models trained on English WDC datasets. \resinfo}
        \label{tab:wdc_comparison}
    	\centering
    	\begin{tabular}{cccccc}
    	\toprule
    		dataset & dataset & \multirow{2}{*}{mBERT} & \multirow{2}{*}{XLM-RoBERTa} & Ditto & WDC-Deepmatcher \\
    		type & size & & & (reported in \cite{ditto}) & (reported in \cite{WDCwwwv2}) \\
    	\midrule
    	\multirow{4}{*}{Cameras} 	& small  & $\mathbf{82.13(\pm 4.70)}$ & $81.96(\pm 7.75)$ & 80.89 & 68.59 \\
    							 	& medium & $87.86(\pm 2.04)$ & $\mathbf{88.11(\pm 4.22)}$ & 88.09 & 76.53 \\
    							 	& large  & $90.88(\pm 2.28)$ & $\mathbf{92.36(\pm 0.76)}$ & 91.23 & 87.19 \\
    							 	& xlarge & - 	 & - 	 & \textbf{93.78} & 89.21 \\
    	\midrule
    	\multirow{4}{*}{Computers} 	& small  & $\mathbf{86.43(\pm 3.69)}$ & $81.10(\pm 13.40)$ & 80.76 & 70.55 \\
    							   	& medium & $\mathbf{90.13(\pm 1.89)}$ & $88.69(\pm 2.19)$ & 88.62 & 77.82 \\
    							 	& large  & $92.48(\pm 2.33)$ & $\mathbf{93.71(\pm 0.77)}$  & 91.70 & 89.55 \\
    							 	& xlarge & - 	 & - 	 & \textbf{95.45} & 90.80 \\
    	\midrule
    	\multirow{4}{*}{Shoes} 		& small  & $\mathbf{79.20(\pm 7.89)}$ & $74.98(\pm 13.36)$ & 75.89 & 73.86 \\
    							 	& medium & $\mathbf{84.11(\pm 3.40)}$ & $81.30(\pm 8.21)$ & 82.66 & 79.48 \\
    							 	& large  & $90.28(\pm 2.36)$ & $\mathbf{91.26(\pm 2.09)}$ &  88.07 & 90.39 \\
    							 	& xlarge  & - 	 & - 	 & 90.10 & \textbf{92.61} \\
    	\midrule
    	\multirow{4}{*}{Watches} 	& small  & $\mathbf{87.31(\pm 1.64)}$ & $83.78(\pm 4.38)$ & 85.12 & 66.32 \\
    							 	& medium & $\mathbf{91.17(\pm 4.21)}$ & $89.50(\pm 3.69)$ & 91.12 & 79.31 \\
    							 	& large  & $93.52(\pm 2.63)$ & $93.62(\pm 0.67)$ & \textbf{95.69} & 91.28 \\
    							 	& xlarge  & - 	 & - 	 & \textbf{96.53} & 93.45 \\
    	\bottomrule
    	\end{tabular}
    \end{table*}

\subsection{Experiments Time} \label{s:experiment_time}

\begin{figure}[hpt]
        \centering
        \includegraphics[width=0.5\textwidth]{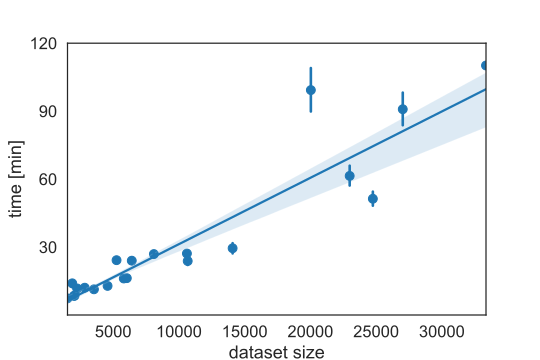}
        \caption{Model fit time vs dataset size. Number of epochs: 10.}
        \label{fig:fit_time}
    \end{figure}

We carried out only four experiments for each dataset. For this reason, to get a standard error with a 95\% confidence level, we had to use t-Student distribution as the number of samples is meagre. As a result, we had to multiply the standard deviation by $\approx 3.18$. This result could be improved by repeating experiments multiple times. After 20 experiments, correction from the t-Student distribution is lowered to $\approx 2.093$. 
Figure~\ref{fig:fit_time} presents the relation between dataset size and time required to train a model. 
    
\section{Summary \& Conclusions} \label{s:conclusions}

This work dealt with the product matching problem. In particular, we thoroughly prepared a new set in Polish and investigated the possibility of using multilingual Transformer models. We compared the models' results trained on the English dataset with other published results. We also prepared the baseline results for the Polish dataset.

Contrary to the previous works in which models were adapted only to the English language, we employed pre-trained multilingual Transformer models in this study. These models achieved similar effectiveness to the other published state-of-the-art solutions utilizing only one-language English models. This work also introduces the first open dataset on the product matching problem in Polish. Furthermore, the obtained data were processed and reduced to a format compatible with the English WDC dataset used to evaluate the product matching task models. The baseline for the newly created dataset was the same as the pre-trained models that previously served as the basis for the English-language dataset. The obtained results showed that it is possible to use the same models for Polish and English data. Hopefully, similar results are also possible with other languages. Thus, it opens up new possibilities for using transfer learning to solve product matching for different languages and shows that it is possible to use pre-trained models instead of preparing models from scratch.

\section*{Acknowledgments} 
We want to thank Maciej Wasiak, Adam Wawrzeńczyk, Marcin Zakrzewski, Karol Ulanowski, Tomasz Zalewski,Jakub Waszkiewicz, Michał Urawski, working under Anna Wróblewska’s guidence at Natural Language Processing course at Warsaw University of Technology, for their valuable contribution into preliminary version of our new dataset and preliminary analysis and baselines, which we modified further.

The research was funded by the Centre for Priority Research Area Artificial Intelligence and Robotics of Warsaw University of Technology within the Excellence Initiative: Research University (IDUB) programme (grant no 1820/27/Z01/POB2/2021).

The work was supported by the Faculty of Physics and Applied Computer Science AGH UST statutory tasks within the subsidy of MEiN.

\FloatBarrier
%\clearpage %TODO: if we want figures before all the references

\bibliographystyle{unsrtnat}
\bibliography{references}  %%% Uncomment this line and comment out the ``thebibliography'' section below to use the external .bib file (using bibtex) .

\end{document}